%% file: main.tex
\pdfoutput=1

\documentclass[11pt]{article}

\usepackage[]{ACL2023}

\usepackage{times}
\usepackage{latexsym}

\usepackage[T1]{fontenc}

\usepackage[utf8]{inputenc}

\usepackage{microtype}

\usepackage{inconsolata}

\usepackage{graphicx}
\usepackage{subcaption}
\usepackage{pifont}
\usepackage{xcolor}
\usepackage{tikz}
\usepackage{amsmath} 
\usepackage{amssymb}
\usepackage{booktabs} 

\definecolor{lightcoral}{RGB}{240,128,128}
\definecolor{darkred}{RGB}{139,0,0}
\definecolor{Aqua}{RGB}{0,255,255}

\title{Unveiling the Secrets of Engaging Conversations: Factors that Keep Users Hooked on Role-Playing Dialog Agents}

\author{\textbf{Shuai Zhang}$^{1,2}$\thanks{\ \ Equal Contribution.}, \textbf{Yu Lu}$^{2}$\footnotemark[1], \textbf{Junwen Liu}$^{3}$ \\ \textbf{Jia Yu}$^{1,2}$, \textbf{Huachuan Qiu}$^{1,2}$, \textbf{Yuming Yan}$^{3}$,  \textbf{Zhenzhong Lan}$^{2}$\thanks{\ \ Corresponding Author.} \\
$^{1}$ Zhejiang University $^{2}$ School of Engineering, Westlake University $^{3}$ Scietrain \\
\texttt{zhangshuai@westlake.edu.cn} \\
}

\begin{document}
\maketitle
\begin{abstract}

With the growing humanlike nature of dialog agents, people are now engaging in extended conversations that can stretch from brief moments to substantial periods of time. Understanding the factors that contribute to sustaining these interactions is crucial, yet existing studies primarily focusing on short-term simulations that rarely explore such prolonged and real conversations.

In this paper, we investigate the factors influencing retention rates in real interactions with role-playing models. By analyzing a large dataset of interactions between real users and thousands of characters, we systematically examine multiple factors and assess their impact on user retention rate. Surprisingly, we find that the degree to which the bot embodies the roles it plays has limited influence on retention rates, while the length of each turn it speaks significantly affects retention rates. This study sheds light on the critical aspects of user engagement with role-playing models and provides valuable insights for future improvements in the development of large language models for role-playing purposes.

\end{abstract}

\input{intro}
\input{related_work}
\input{methods}
\input{experimental_results}

\section{Conclusion}
In this study, we investigated the factors that affect user retention rates in a role-playing model. We analyzed various potential factors and compared them systematically between strong and weak model pairs selected from an A/B testing pool. Our findings shed light on the previously overlooked positive influence of non-verbal description and length. Additionally, contrary to our initial intuition, repetition was found to have a positive impact on user retention rates. We observed that factors such as human-likeness and personality consistency aligned with expectations, while diversity, fact consistency, empathy, and proactivity did not significantly affect user interaction rates. These insights provide valuable information for future research aimed at enhancing role-playing models. However, our analysis may not have considered all relevant variables, generalization should be done cautiously. 

\section{Limitation}
While our work has made significant contributions, it is important to acknowledge the following limitations:

(i) Our approach to identifying factors influencing user retention rates relies on statistical correlations. However, it is crucial to note that correlation does not necessarily imply causation between the factors and interaction preferences. Further research is needed to establish causal relationships.

(ii) The quantification and calculation of factors are subject to the limitations of the tools we utilized, such as representation models or APIs. As these tools continue to evolve and improve, our results may benefit from their advancements.

(iii) User retention rates are influenced by a multitude of factors working in combination. However, our study only considered these factors individually. To gain a comprehensive understanding, future work should explore the combined effects of these factors.

\section*{Ethics Statement}
The data used in this research is derived from real user interactions with the role-playing model. Users have provided consent for their data to be used for scientific research under our terms of service. Importantly, the data we have utilized does not contain any personal details, ensuring the privacy and confidentiality of our users.

\bibliography{main}
\bibliographystyle{acl_natbib}

\appendix

\input{appendix}
\end{document}

%% file: intro.tex
\section{Introduction}

As dialog agents continue to advance in their humanlike capabilities, individuals are increasingly engaging in extended conversations with these agents that can span hours, days, or even months \cite{irvine2023rewarding,buse2023social}. These interactions with role-playing chatbots have become a significant aspect of people's daily lives, providing companionship, entertainment, and even therapeutic benefits \cite{buse2023social}. Understanding the factors that contribute to sustaining these interactions is crucial for both researchers and developers, as it can help improve the design and effectiveness of these conversational agents \cite{lee2022evaluating,shen2023roleeval}.

While existing studies have provided valuable insights into user interactions with dialog agents, they have primarily focused on short-term simulations \cite{wang2023rolellm,zhou2023characterglm,tu2024charactereval,fable2023showrunner,chen2023large}. These studies have examined factors such as coherence, fluency, and relevance to assess user engagement. However, the dynamics and challenges associated with long-term interactions, where real users engage in ongoing conversations with role-playing models, have received less attention. A problem with these simulated conversation and evaluation \cite{shen2023roleeval} is that they may not truly reflect what real users care about. 

In this paper, we aim to bridge this gap by investigating the factors that influence user retention rates in real interactions with role-playing models. To accomplish this goal, we analyze a large dataset comprising interactions between real users and thousands of characters. This dataset captures the richness and complexity of extended conversations, providing a comprehensive view of user engagement with role-playing chatbots.

Our analysis focuses on multiple factors that may influence user retention rates. We consider factors such as the alignment of the model's responses with the roles it plays, the length of each turn, the human-likeness, and the diversity of the words used, etc. By systematically examining these factors, we aim to identify the key drivers that contribute to sustaining interactions with role-playing models over extended periods.

Surprisingly, our findings reveal that the degree to which the bot embodies the roles it plays has limited influence on user retention rates. Instead, we find that the length of each turn the bot speaks plays a crucial role in maintaining user interest. Longer turns tend to lead to higher retention rates, suggesting that users prefer more substantial and in-depth responses from the bot.

These findings have important implications for the development and design of large language models for role-playing purposes. By prioritizing the length and depth of model responses, developers can enhance user engagement and encourage longer and more meaningful conversations. Additionally, our study provides valuable insights into the dynamics of user engagement with role-playing models, contributing to the broader understanding of human-computer interactions and the potential applications of conversational agents in various domains.

Overall, this paper sheds light on the critical aspects of user engagement in extended conversations with role-playing models. By analyzing a large dataset and systematically examining various factors, we uncover the limited influence of role embodiment, the significant impact of turn length on user retention rates and other important factors. These findings offer valuable guidance for researchers and developers seeking to improve the effectiveness and long-term engagement of role-playing chatbots and lay the foundation for further advancements in this field.

%% file: related_work.tex
\section{Related work}

Our work aims to investigate the factors influencing user retention rate in role-playing models. This investigation is closely tied to the research on role-playing model evaluation. Existing approaches to model evaluation mainly rely on role-playing data that is not derived from real user interactions. Instead, these data often come from character-based materials such as movies, novels, scripts, or Wikipedia entries \cite{li2023chatharuhi,tu2024charactereval,salemi2023lamp} for two-party conversation or even multi-part conversation\cite{wei2023multi}. Some approaches use synthetic data generated from ChatGPT \cite{wang2023rolellm}, while others utilize conversations with a limited number of annotators or well-designed question data \cite{zhou2023characterglm, shen2023roleeval, shao2023character, wang2023does,park2023generative}. In contrast, our evaluation is based on conversation data obtained from a large number of real users interacting with the role-playing model.

Furthermore, existing model evaluation strategies primarily focus on the alignment of the model's response with predefined labels or a third-person perspective \cite{wang2023rolellm, shen2023roleeval,li2023chatharuhi, zhou2023characterglm}. These evaluations primarily assess the model's ability to generate accurate or contextually appropriate responses, often based on predefined criteria or external references. However, these evaluation may not reflect the needs of real users. In contract, we consider these preferences from a first-person perspective, directly from the users themselves.

%% file: methods.tex
\section{Methods}
\begin{figure*}[h]
  \centering
  \includegraphics[width=\linewidth]{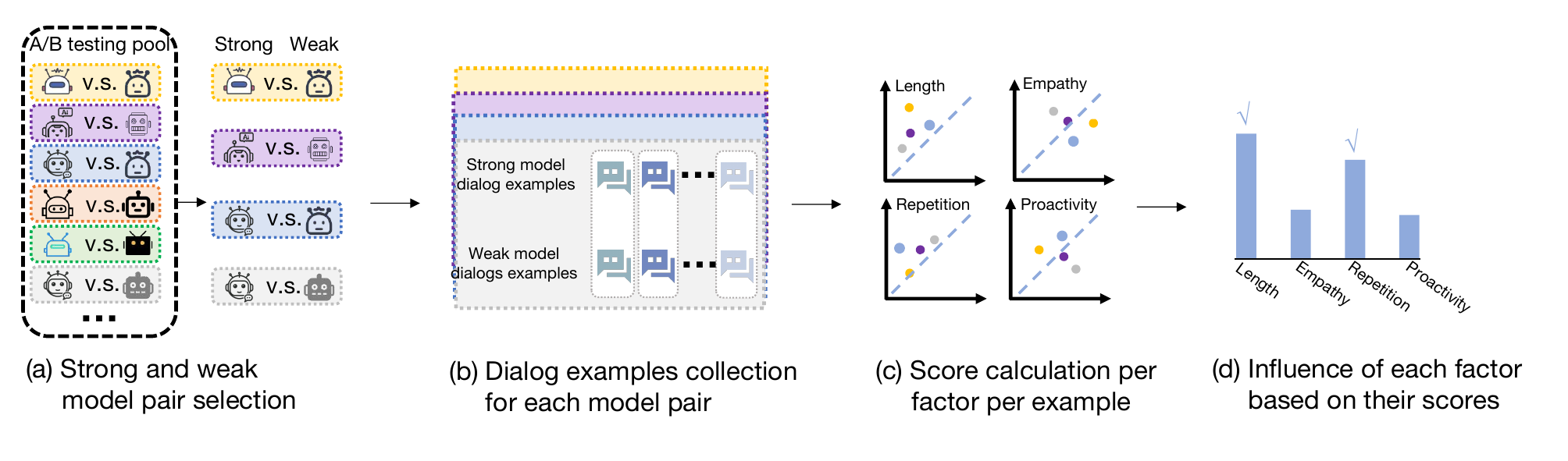}
  \caption{Our method follows a pipeline that begins by identifying pairs of (strong and weak) models with significantly different retention rates through A/B testing. From these pairs, we then sample dialog examples and utilize them to calculate scores for various factors. Finally, we quantify the impact of these factors and their significance by analyzing the calculated scores. }
  \label{fig:method_pipeline}
\end{figure*}

As shown in Figure \ref{fig:method_pipeline}, we first engage a considerable number of users in role-playing interactions with a variety of role-playing models through A/B testing. To ensure that the comparison of each model pair is statistically significant, we make sure that each model has at least 140 new users for each testing day and last for at least 7 days. In these 7 days, the user retention rate of the strong model has to exceed that of the weak model for at least 6 days to qualify as a valid pair. At the end of our month's testing periods, we have collected 53 valid model pairs.

Once we have identified these model pairs, we randomly select dialog data from these models as a valuable resource for uncovering potential influences on user interaction preferences. The sampled dialog data provides us with insights into the patterns and dynamics of user engagement. Specifically, we sample $N$ = 1000 dialogs for each model and together, we have 10600 dialogs for all 53 model pairs.

Next, we meticulously enumerate and quantify potential factors within the role-playing model that could impact user interaction preferences. These factors are computed and analyzed using the sampled dialog data for each model. By examining these factors, we gain a deeper understanding of the underlying mechanisms that drive user retention. We will provide the details of how we calculate these factors in the following section. 

Finally, we identify influential factors by scrutinizing whether their values consistently exhibit significant differences across multiple pairs of models with varying retention rates. This analysis allows us to pinpoint the specific factors that have a significant impact on user retention and interaction preferences.

\subsection{Factors Definition}

Together, we thoroughly analyze 9 factors that potentially impact user retention rates. These factors are derived from extensive research conducted by \citet{wang2023rolellm} and \citet{shen2023roleeval}, supplemented by our own insights gained from reviewing the collected dialog samples. In this description, we will provide a comprehensive rationale for selecting these factors and elucidate the methodology employed to calculate their scores.

For the first four factors, we employ straightforward methods to determine their scores. However, for other factors such as human-likeness, we rely on the advanced capabilities of GPT-4 to assess whether the models exhibit human-like behavior. The simplified versions of the prompts used for GPT-4 evaluation are presented in Table \ref{table:prompt_template_simplifed}, while the complete versions can be found in Table \ref{table:prompt_template} in the appendix. As it is expensive to call GPT-4 API, we segment the dialogs into consecutive slices, each containing five turns. and randomly sample a subset of $M$ = 100 slices for each GPT-4 evaluated factors.   

\noindent\textbf{Length}: We are interested in determining the average length of each turn in order to assess the amount of information contained in each response. To calculate this, we utilize the mean word count across all utterances in the sampled dialogs, as defined by the following equation:

\begin{equation}
\frac{1}{\sum_{n=1}^N T_n}\sum_{n=1}^{N}\sum_{t=1}^{T_n}l^t_n \nonumber
\end{equation}
In this equation, $N$ represents the number of dialogs, while $T_n$ denotes the number of utterances in the $n$-th dialog. The variable $l^t_n$ corresponds to the number of words in the $t$-th utterance.

\noindent\textbf{Diversity}: Diversity in the model's response reflects the richness of vocabulary used. A higher word diversity implies more flexibility in expression, potentially conveying additional information and preventing the interaction from becoming monotonous. This, in turn, positively influences user interaction preferences.

To calculate the mean diversity across all utterances in the sampled dialogs, we employ the following equation:

\begin{equation}
 \frac{1}{\sum_{n=1}^N T_n}\sum_{n=1}^{N}\sum_{t=1}^{T_n}\prod_{n=2}^4d^t_n \nonumber
\end{equation}
where, $d^t_n$ is calculated as the ratio of the number of unique n-grams to the total number of n-grams in the utterance, after removing duplicates, as specified in \citeauthor{su2022contrastive}.

\noindent\textbf{Repetition}: Intuitively, semantic repetition in consecutive utterances of a model can make the interaction monotonous and inefficient, negatively impacting user interaction preferences. To measure this, we calculate the proportion of semantically repetitive adjacent utterances in each dialog. The mean proportion is computed using the following equation:

\begin{equation}
    \frac{1}{N}\sum_{n=1}^N \frac{1}{T_n-1}\sum_{t=1}^{T_n-1}\text{I}_{\text{repetition}}(u_n^t, u_n^{t+1}) \nonumber
\end{equation}
In this equation, $\text{I}_{\text{repetition}}(u_n^t, u_n^{t+1})$ is an indicator that determines whether the adjacent utterances of the model, $u_n^t$ and $u_n^{t+1}$, are semantically repetitive. We assess the semantic similarity between two sentences by computing the cosine similarity of their Jina Embeddings \cite{gunther2023jina}. If the similarity exceeds 0.95, we consider the sentences as semantically repeated.

\noindent\textbf{Non-verbal Description}: The occurrence of non-verbal descriptions in the model's response provides detailed information about environmental changes and the physical and psychological reactions of characters in the conversation. For instance, actions like "I walk towards you" create a vivid experience of how the bot interacts with users. These non-verbal descriptions are more common in role-playing conversations compared to other types of dialog. We are interested in investigating whether these descriptions influence user interaction preferences. To measure their frequency, we can count the number of utterances that contain non-verbal descriptions. Mathematically, this can be represented as:

\begin{equation}
\frac{1}{\sum_{n=1}^N T_n}\sum_{n=1}^N\sum_{t=1}^{T_n} \text{I}_{\text{non-verbal}}(u_n^t) \nonumber 
\end{equation}
Here, $\text{I}_{\text{non-verbal}}(u_n^t)$ is an indicator that determines whether the model's utterance $u_n^t$ includes a non-verbal description.

\noindent\textbf{Human-likeness}: Characters in our system exhibit human-like behavior by expressing their personal preferences, feelings, and opinions. This anthropomorphic aspect creates a sense of realism, making users feel as though they are interacting with real-world characters. This can greatly enhance the user experience and preference for interaction. To evaluate whether the dialog exhibit human-likeness, we employ GPT-4 and utilize the prompt template shown in Table \ref{table:prompt_template_simplifed} for assessment. The score is thus calculated  as the  the mean frequency of the model expressing the human-likeness using the subsequent equation: 
\begin{equation}
 \frac{1}{M} \sum_{m=1}^M\text{I}_{\text{human-likeness}}(C_m) \nonumber 
\end{equation}
where $M$ is the number of sampled slices and $\text{I}_{\text{human-likeness}}(C_m)$ is an indicator that signifies whether the utterance of the model from the chunked sub-dialog $C_m$ reflects human-likeness.\\
\noindent\textbf{Fact Consistency}: We aim to assess the coherence between the model's statements and the facts of personal attributes of the character it represents. These facts encompass intrinsic traits like birthdate, gender, family members, as well as acquired experiences such as specific times, locations, and significant events in the character's life. Naturally, the portrayal of a character becomes more lifelike when the model's statements harmonize effectively with the character's personal attributes. To ensure this consistency, we employ the GPT-4 API to verify whether the utterance entails, contradict, or is independent of the facts in the character profile. The overall score of fact consistency is calcuated as following equation: 

\begin{equation}
    \sum_{m=1}^M s(C_m, P_m^f) \nonumber 
\end{equation}
where 
\begin{equation}
s(C_m, P_m)=\begin{cases}
1, \text{if } C_m \text{ entails } P_m^f \\
0, \text{if } C_m \text{ is independent of } P_m^f \\
-1, \text{if } C_m \text{ contradicts } P_m^f
\end{cases}
 \nonumber 
\end{equation}
where the $C_m$ is the $m$-th chunked sub-dialog and $P^f_m$ is the fact part of the profile information. 

\noindent\textbf{Personality Consistency}: Personality consistency shares similarities with fact consistency, but it focuses on evaluating different elements. In our evaluation, we consider personality elements that contribute to the character's unique identity. These elements encompass various aspects, including catchphrases, hobbies, personality traits, and values, as outlined in Table \ref{table:prompt_template_simplifed}. By examining the consistency of these personality elements, we can assess how well the model portrays the character's distinctive traits and maintains coherence with their established personality throughout interactions. The formal score calculation is as following: 

\begin{equation}
    \sum_{m=1}^M s(C_m, P_m^p) \nonumber 
\end{equation}
where 
\begin{equation}
s(C_m, P_m^p)=\begin{cases}
1, \text{if } C_m \text{ entails } P_m^p \\
0, \text{if } C_m \text{ is independent of } P_m^p \\
-1, \text{if } C_m \text{ contradicts } P_m^p
\end{cases}
 \nonumber 
\end{equation}
where the $C_m$ is the $m$-th chunked sub-dialog and $P^p_m$ is the personality part of the profile information. 

\noindent\textbf{Empathy}: The role-playing model demonstrates its capacity to recognize, comprehend, and acknowledge the user's emotions, responding with an appropriate tone and language that offers supportive or helpful content. This ability is believed to assist users in addressing their emotional concerns, fostering emotional connections, and ultimately influencing their interaction preferences in a positive manner. By effectively understanding and responding to the user's emotions, the model aims to provide a more personalized and empathetic experience, enhancing the overall user satisfaction and engagement.

We compute the sum of empathy scores for all sampled chunked sub-dialogs according to the following equation:
\begin{equation}
\sum_{m=1}^M s(C_m) \nonumber
\end{equation}
The function $s(C_m)$ is defined as:
\begin{equation}
s(C_m)=\begin{cases}
0, & \text{for NoEmotion}\\
1, & \text{for EmotionWithEmpathy}\\
-1, & \text{for EmotionWithoutEmpathy}\\
\end{cases} \nonumber
\end{equation}
Here, 'NoEmotion' indicates that no emotion of the user is detected in $C_m$. 'EmotionWithEmpathy' denotes scenarios where the user exhibits emotion and the character displays empathy in $C_m$. 'EmotionWithoutEmpathy', signifies instances where the user is emotional but the character does not exhibit empathy in $C_m$.

\noindent\textbf{Proactivity}: The role-playing model possesses the valuable skill of proactive response, enabling the character to actively engage and interact with the user. This capability has the potential to stimulate user engagement within the conversation and is believed to enhance user interaction preferences. To simplify our analysis, we focus on three specific types of proactive behavior: asking for clarification, eliciting user preferences, and engaging in target-guided dialog. For a comprehensive understanding of these three activities, please refer to \cite{deng2023survey} and Table \ref{table:prompt_template} in the Appendix. By incorporating these proactive elements, the model aims to foster a more dynamic and engaging conversation, ultimately leading to an improved user experience.

We calculate the frequency of the model's proactive responses using the following equation:
\begin{equation}
\frac{1}{M}\sum_{m=1}^M\text{I}_{\text{proactivity}}(C_m) \nonumber
\end{equation}
where $\text{I}_{\text{proactivity}}(C_m)$ indicates whether the model responds proactively to the user in the chunked sub-dialog $C_m$.
\label{sec:appendix_1}

\begin{table*}
\centering
\begin{tabular}{p{0.12\linewidth}p{0.82\linewidth}}
\hline
\textbf{Factors} & \textbf{Simplified version of prompt given to GPT-4}\\
\hline
Human-likeness &
... Evaluate the subjectivity of the character that the assistant is portraying. This involves determining whether the assistant's responses express the personal preferences, feelings, or opinions of the character they are playing...\\
\hline
Fact or Personality Consistency &
... The assistant is portraying a character named \$character\_name, and the character's profile will be provided below. Your task is to assess whether the assistant's responses are consistent with the character's profile in terms of fact consistency and personality consistency.
\textbf{Fact consistency:} Determine if the assistant's statements clearly align with or contradict the factual aspects of the character, including innate facts that the character is born with such as birthday, gender, era, family members, etc, and facts that the character acquires throughout their life, such as the time, place, related people, and events of the character's first job, the positions they've held in their career, and other significant life experiences, etc.
\textbf{Personality consistency:} Evaluate if the assistant's statements obviously match or contradict the character's personality. Personality elements include aspects such as the character's catchphrases, hobbies, personality traits, and values, etc.  ... \\
\hline
Empathy &
... Your task is to assess the empathy exhibited by the character that the assistant is portraying. To accomplish this, begin by determining whether the user is explicitly expressing positive or negative emotions. Subsequently, evaluate whether the character portrayed by the assistant responds empathetically to the user's emotions. ... \\
\hline
Proactivity &
... Evaluate if the assistant's responses are proactive, instead of merely passively responding to the user. There are three types of proactivity: Asking for Clarification: ...
User Preference Elicitation: ...
Target-guided Dialog:...\\
\hline
\end{tabular}
\caption{\label{table:prompt_template_simplifed} Prompt template for factors that require the use of GPT-4 API.
}
\end{table*}

\subsection{Significance Test}
To assess the impact of each factor on user rentention rates, we compare the scores between strong and weak models for all pairs. Factors showing significant score differences are considered influential for user retention rates.

\textbf{Mean Score Difference}: We calculate the score difference $d_i = s_i - w_i$ for each factor in both strong and weak models, where $s_i$ and $w_i$ represent the scores of the factor for the $i$-th group's strong and weak models, respectively. The mean score difference $d = \frac{1}{n}\sum_{i=1}^n d_i$ across all groups quantifies the average difference of the factor. Intuitively, a higher positive (negative) mean difference indicates a greater proportion of groups with strong models having a positive (negative) difference compared to weak models.

\textbf{Significance of the Difference}: We employ permutation tests to determine the significance of the difference. By randomly swapping scores between pairs of strong and weak models, we calculate a new average difference score $\hat{d} = \frac{1}{n}\sum_{i=1}^n \hat{d}_i$, where $\hat{d} = (-1)^{p_i}d_i$, and $p_i \stackrel{\text{i.i.d.}}{\sim} \text{Bernoulli}(0.5)$. According to the Lyapunov Central Limit Theorem, when the number of model pairs is sufficiently large, $\hat{d}$ approximates a normal distribution, denoted as $\hat{d} \sim \mathcal{N}(\mu, \sigma^2)$. Subsequently, we compute the z-score as $z = \frac{d - \mu}{\sigma} \sim \mathcal{N}(0, 1)$. A positive z-score suggests a positive influence on user retention rates, while a negative z-score indicates a negative influence. The tail probability of the observed z-score under the normal distribution is used to determine the significance. Intuitively, a larger positive (negative) mean difference score ranks higher (lower) among all permuted mean difference scores, indicating a larger positive (negative) z-score, a smaller p-value, and a more significant positive (negative) influence of the factor on user retention rate. A p-value less than 0.05 is typically considered to be statistically significant. We perform 100,000 random swaps to obtain 100,000 $\hat{d}$ values and estimate $\mu$ and $\sigma$ by calculating the mean and standard deviation, respectively.

%% file: experimental_results.tex
\section{Experimental results}
\subsection{Setup}

We carefully selected a total of 53 pairs of strong and weak models from A/B testing pools during a four-month period, specifically from July 25, 2023, to November 25, 2023.
These models were trained by researchers who actively contribute to a publicly accessible platform for role-play conversation.
The key differences among these models lie in their foundation models (LLaMA \cite{touvron2023llama,touvron2023llamav2} or Mistral \cite{jiang2023mistral}), as well as their continuous pre-training and fine-tuning methods. It's worth noting that we have obtained permission to publish some of these models along with the paper.

We conducted an analysis on the impact of model factors on user retention rates by randomly selecting a sample of $N=1000$ dialogs from each model. The statistics of the sampled conversations are presented in Table \ref{tab:data_stat}. Notably, our dataset comprises over 37,000 users engaged in conversations totaling more than 8,000 characters. This dataset stands out as the largest realistic conversation dataset in terms of both the number of users and the amount of text involved.

It is worth mentioning that the average utterance per speaker is relatively low. This observation can be attributed to the fact that a significant portion of conversations do not delve deeply. Many users initiate a conversation, engage briefly, and then leave. However, when conversations do become more involved, they can extend to thousands of turns within just a few days.

Comparing the average length of utterances between users and characters, we observe that characters tend to contribute significantly more to the conversation. This discrepancy can be attributed to the fact that many users exhibit limited conversational skills, often expressing themselves with only a few words or phrases. It is crucial for bots to compensate for this by generating substantial dialog, even in situations where users struggle to initiate or sustain a conversation.

\begin{table}[t]
\scalebox{0.8}{
\begin{tabular}{@{}cccc@{}}
\toprule
\textbf{Category}              & \textbf{Total}     & \textbf{User} & \textbf{Character} \\ \midrule
\#Dialogs                      & $106,000$          & -             & -                  \\
\#Speakers                     & $45,867$           & $37,752$      & $8,115$            \\
\#Avg. utterance per speaker   & $11.47$            & $11.03$       & $11.91$            \\
\#Avg. length per utterance    & $37.90$            & $10.86$       & $62.39$            \\ 
\bottomrule 
\end{tabular}}
\caption{Statistics of the sampled conversation. }
\label{tab:data_stat}
\end{table}

\begin{figure}[h]
  \centering
  \includegraphics[width=\linewidth]{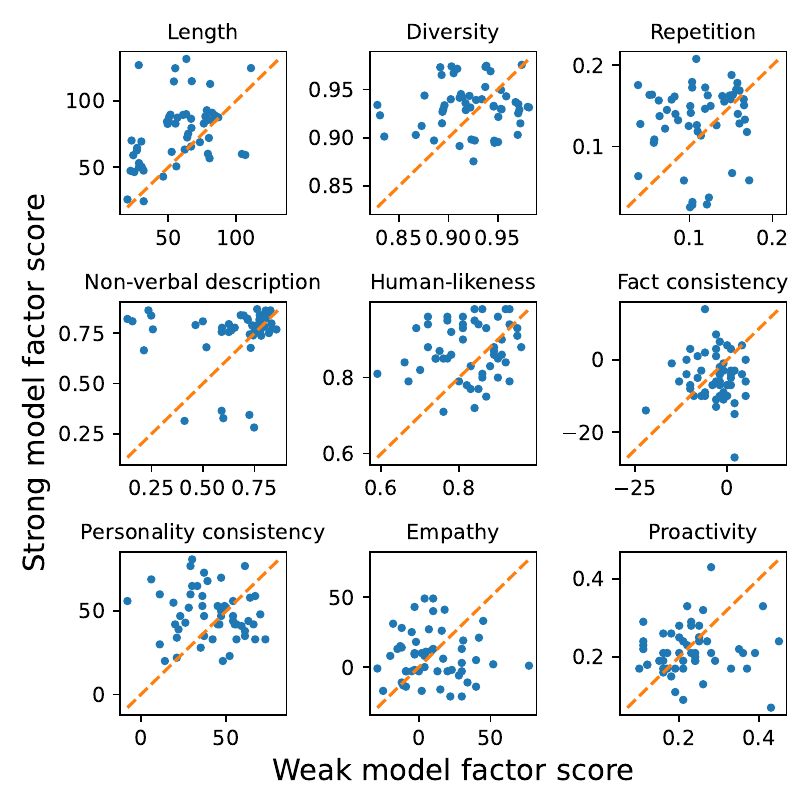}
  \caption{Comparison of factor scores between strong and weak models is visualized in a grid of sub-figures. Each point in the grid represents a pair of strong and weak models, with the x-axis and y-axis denoting the factor scores of the strong and weak models, respectively. A dotted line within each sub-figure indicates points with identical factor scores between the model pairs. More points located towards the top-left (bottom-right), away from the dotted line, suggest a stronger positive (negative) influence on the retention rates. }
  \label{fig:feature_score}
\end{figure}

\begin{figure}[h]
  \centering
  \includegraphics[width=\linewidth]{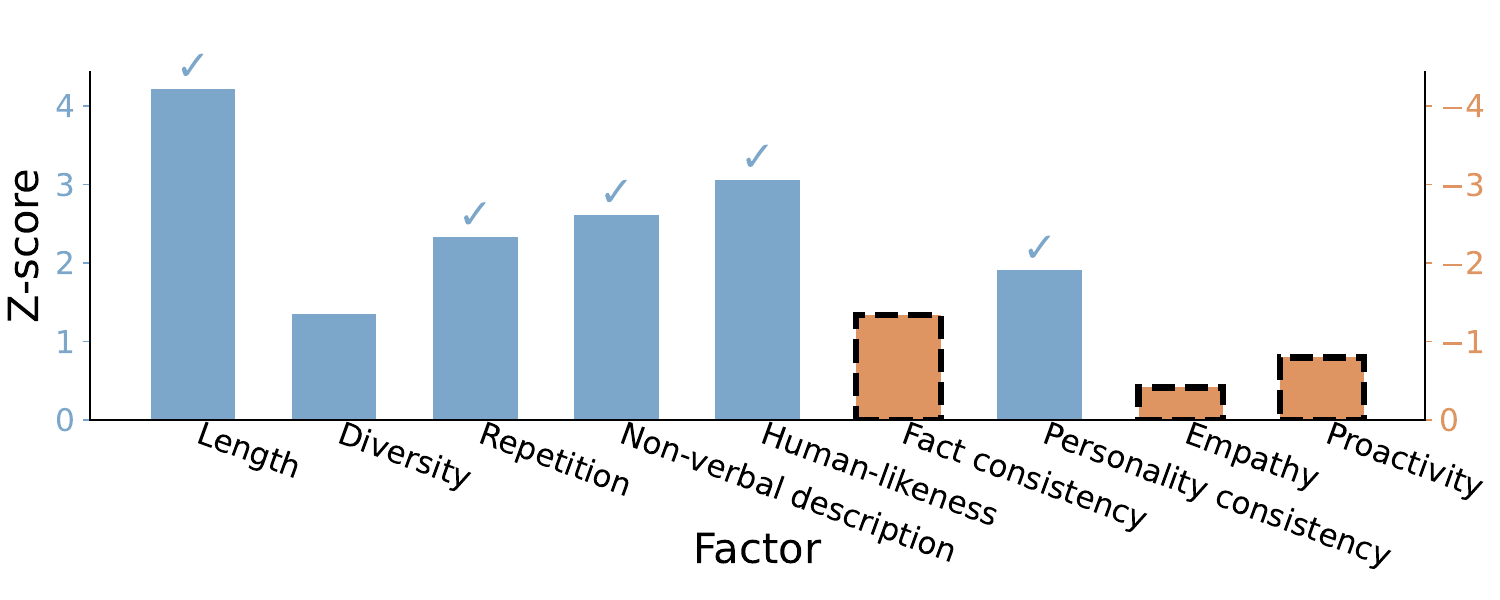}
  \caption{This figure shows the z-scores of each factor. \textcolor{blue}{Blue} bars represent factors with positive z-scores, while \textcolor{orange}{orange} bars with dashed outlines represent factors with negative z-scores. The presence of a check marker on top of the bars indicates that these factors have a statistically significant influence on retention rates (p-value < 0.05). }
  \label{fig:significance}
\end{figure}

\subsection{Overall results}
 Figure \ref{fig:feature_score} shows the comparison of score of each factor between the strong and weak models for all pairs. Figure \ref{fig:significance} shows the significance of each factor's influence on user retention.

From Figure \ref{fig:feature_score} and \ref{fig:significance}, we observe that only length, repetition, non-verbal description, human-likeness, and personality consistency have a significant impact on user retention (p-value<0.05). Conversely, diversity, fact consistency, empathy, and proactivity do not appear to influence user retention. In the subsequent section, we will provide a detailed analysis of each individual factor.

\subsection{Factor Analysis}

\noindent\textbf{Length}. Figure \ref{fig:feature_score} clearly demonstrates that the average utterance length has the most substantial impact on the retention rate. It is noteworthy that many of the strong models tend to speak longer than their weak counterparts, with some strong models having an average utterance length exceeding 100 words.

While length is a somewhat broad measure, it generally indicates the amount of information conveyed in the response. This correlation provides the model with a valuable opportunity to express comprehensive and detailed information, encompassing the character's verbal expressions and associated non-verbal context.

During a conversation, longer responses have the advantage of effectively conveying information with fewer conversational turns. This efficiency allows users to quickly grasp and immerse themselves in the role-playing world with less effort. Furthermore, it is important for the bot to speak at length because users are actively co-creating stories with the bots. In this context, the bot serves as a storyteller and needs to provide intricate details and continue the narrative (See Table \ref{table:length_example} in the Appendix for an illustrative example).

\noindent\textbf{Diversity}. We did not observe a significant influence of the model's response word diversity on user retention rates. A plausible explanation could be that the level of diversity across all models is sufficiently high (Figure \ref{fig:feature_score}), such that variations in diversity among different models do not notably affect the users' experiences. The impact of it may be overshadowed by other factors.

\noindent\textbf{Repetition}. Contrary to our initial expectations, our findings indicate that frequent semantic repetition within consecutive responses from the models can significantly enhance user retention rates (Figure \ref{fig:significance}). Upon closer inspection of these repetitions, we discovered that much of the semantic repetition occurs in non-verbal segments of the complete utterance, while verbal words remain distinct, leading to repetition of the entire sentence. As a result, we posit that this repetition is instrumental in ensuring the non-verbal context evolves in a consistent and continuous manner along with the flow of the conversation, thereby maintaining the coherence of the role-playing world (see an example in Table \ref{tabel:repetition_example} in the Appendix).

\noindent\textbf{Non-verbal Description}. 
Figure \ref{fig:significance} suggests that frequent utilization of non-verbal descriptions can enhance a role-playing model's capacity to increase user retention. The effectiveness of non-verbal descriptions lies in their ability to foster a rich and immersive context, which promotes user to engage in joyful and sincere communication within the role-playing environment, thereby intensifying their sense of enjoyment and belonging. On one hand, it enriches the interaction elements by incorporating characters' behaviors and the surrounding environment, not merely a character's verbal words. This approach renders the role-playing world physically visible, enabling users to interact not only with the characters but also with the characters' world. As a result, it provides huge space for users to explore and manipulate, thereby heightening their enjoyment (See an example in Table \ref{table:non_verbal_example_1} in the Appendix). On the other hand, non-verbal description offers insights into characters' genuine feelings and thoughts, making the role-playing world psychologically perceptible. This serves as a crucial medium for users to establish a connection with the characters, fostering a sense of belonging (an example is given in Table \ref{table:non_verbal_example_2} in the Appendix).

\noindent\textbf{Human-likeness}. As expected, human-likeness in the model's responses do significantly enhances user retention rate (Figure \ref{fig:significance}). The model not only displays facts and behaviors of the character during interactions with users but also presents the character's personal feelings, preferences, and opinions. This imbues the character with a sense of unpredictability, akin to a real person with their unique choices, thereby bringing the character to life (See an example in Table \ref{tabel:character_subjectivity_example} in the Appendix).

\noindent\textbf{Fact Consistency}. Our results suggest that the alignment between the model's responses and the assigned character's factual background does not significantly impact user retention rates (Figure \ref{fig:significance}). We've noticed that users often immerse themselves in scenarios pertaining to the character's life. Conversations may touch upon elements of the character's life, yet don't necessarily adhere to the character's pre-established facts. During user-character interaction, 'new facts' often emerge. These may neither coincide with nor contradict the character's existing facts. In dialogs, the character's personality is often more prominently featured than their factual background. Only 14.2\% of all chunked sub-dialogs present the character's personal facts, compared to 87.5\% of all chunked sub-dialogs that showcase the character's personality. We've also observed that any inconsistencies with the facts are often minor (Table \ref{table:fact_consistency_example}). This suggests that users aren't primarily concerned with verifying the character's known facts. Instead, they seem keen on exploring the character's life and crafting new narratives, showing a tolerance for occasional minor factual discrepancies.

\noindent\textbf{Personality Consistency}. As anticipated, our results concur that a model which aligns characters' responses more accurately with their respective personalities can significantly enhance user retention rates (Figure \ref{fig:significance}). Users tend to engage more with characters whose words and behaviors accurately embody their expected characteristics.

\noindent\textbf{Empathy}. From the results, we cannot see that the model's empathy capability significantly influence user retention rates (Figure \ref{fig:significance}). Upon closer inspection, it appears that users may feign emotional responses to align with role-playing scenarios. The characters, in turn, may respond to the user's emotions in multiple ways, not limited to empathetic responses. This dynamic potentially enriches the character's personality and makes their conversations more engaging and animated (an example is shown in Table \ref{table:empathy_example} in the Appendix). 

\noindent\textbf{Proactivity}. Our results suggest that the model's proactivity capability does not significantly influence user retention rates (Figure \ref{fig:significance}). Based on our observations, it appears that users may engage in conversations where the model provides in-depth, relevant, and user-guided responses, with proactive responses not serving as a key factor. This interaction style creates a sense of involvement and attractiveness, making users feel their inputs are valued and understood, which encourages them to continue participating in the conversation (an example is shown in Table \ref{table:proactivity_example} in the Appendix).

%% file: appendix.tex
\newpage
\section{Exemplar dialogs}

\begin{table*}
\centering
\begin{tabular}{p{0.12\linewidth}p{0.88\linewidth}}
\hline
\textbf{Factors} & \textbf{Prompt template}\\
\hline
Human-likeness &
Review the following role-playing dialog. Evaluate the subjectivity of the character that the assistant is portraying. This involves determining whether the assistant's responses express the personal preferences, feelings, or opinions of the character they are playing. Please provide your answer in the following JSON format: \{"subjectivity": "True or False", "explanation": ""\}. \textbackslash n[Start of dialog] \textbackslash n\$dialog\textbackslash n[End of dialog] \\
\hline
Fact Consistency \& Personality Consistency &
Review the following dialogue from a role-playing scenario. The assistant is portraying a character named \$character\_name, and the character's profile will be provided below. Your task is to assess whether the assistant's responses are consistent with the character's profile in terms of fact consistency and personality consistency.
Fact consistency: Determine if the assistant's statements clearly align with or contradict the factual aspects of the character, including innate facts that the character is born with such as birthday, gender, era, family members, etc, and facts that the character acquires throughout their life, such as the time, place, related people, and events of the character's first job, the positions they've held in their career, and other significant life experiences, etc.
Personality consistency: Evaluate if the assistant's statements obviously match or contradict the character's personality. Personality elements include aspects such as the character's catchphrases, hobbies, personality traits, and values, etc. 
You need to identify the points of consistency or contradiction in facts and personality from the dialogue. You need to answer in the following JSON format \{"Consistent Facts": "concise answer and N/A if no consistent facts ", "Contradictory Facts": "concise answer and N/A if no contradictory facts", "Consistent Personality": "concise answer and N/A if no consistent personality", "Contradictory Personality": "concise answer and N/A if no contradictory personality", "explanation": ""\}. Your judgement must be based on clear evidence from the provided dialogue and character profile, without making unfounded assumptions. \textbackslash n[Start of character's profile] \textbackslash n\$profile \textbackslash n[End of character's profile] \textbackslash n[Start of dialogue] \textbackslash n\$dialogue \textbackslash n[End of dialogue] \\
\hline
Empathy &
Review the following role-playing dialog. Your task is to assess the empathy exhibited by the character that the assistant is portraying. To accomplish this, begin by determining whether the user is explicitly expressing positive or negative emotions. Subsequently, evaluate whether the character portrayed by the assistant responds empathetically to the user's emotions. This involves assessing if the character can comprehend and acknowledge the user's emotions, employ suitable tone and language, and offer supportive or helpful content. Your response will fall into one of three scenarios: 1. The user does not display emotions, so the character played by the assistant is not required to respond empathetically; 2. The user displays emotions and the character played by the assistant responds empathetically; 3. The user displays emotions and the character played by the assistant does not respond empathetically. Please provide your assessment in the following JSON format: {"emotion type": "Positive or Negative or Neutral", "empathy": "Yes or No when emotion type is positive or negative, otherwise Not required", "explanation": ""}. \textbackslash n[Start of dialog] \textbackslash n\$dialog \textbackslash n[End of dialog] \\
\hline
Proactivity &
Review the following role-playing dialog. Evaluate if the assistant's responses are proactive, instead of merely passively responding to the user. There are three types of proactivity:
Asking for Clarification: The assistant seeks clarification when the user's input is insufficient, ambiguous, or incorrect.
User Preference Elicitation: The assistant actively inquires about the user's personal preferences.
Target-guided Dialog: The assistant actively creates or controls the conversation.
Please provide your answer in the following JSON format: \{"proactivity": "True or False", "type": "Asking for Clarification or User Preference Elicitation or Target-guided Dialog", "explanation": ""\}. \textbackslash n[Start of dialog] \textbackslash n\$dialog \textbackslash n[End of dialog] \\
\hline
\end{tabular}
\caption{\label{table:prompt_template} Prompt template for factors that require the use of GPT-4 API
}
\end{table*}

\begin{table*}
\centering
\begin{tabular}{p{\linewidth}}
\hline
\textbf{Non verbal description example 1} \\
\hline
ASSISTANT:\textbf{*shrinking in the background*} \\
USER:*i try to sneak through the halls of the school as people call me demon for reasons you dont know* \\
ASSISTANT:\textbf{*as you try to sneak through the hallways, you hear a group of students gossiping about you*} "Did you hear? That new kid is a demon! I heard they came from the deepest and hottest parts of Hell!" \textbf{*one of the students points at you*} "Look! There's the demon now! Stay away from me, demon!" \textbf{*the group backs away in fear*} \\
USER:*i sadden slightly* im not a demon... \\
ASSISTANT:\textbf{*the students continue to back away from you, some even covering their faces in fear*} "Please don't hurt us! We're sorry if we offended you. We didn't know you weren't really a demon." \textbf{*one daring student steps forward*} "So what are you? If you're not a demon, then what are you?" \textbf{*the group surrounds you, curiosity overcoming their fear*} \\
USER:im just a normal dragon! just like everyone else! *as people look at me like im some strange new species instead of a dragon like everyone else i run off* \\
ASSISTANT:\textbf{*you run down the hallway, trying to escape the stares and whispers of the other students. As you turn a corner, you collide into someone, causing them to stumble and fall*} "Hey! Watch where you're going!" \textbf{*you look up to see who you ran into, only to find that it's the most popular girl in your grade. She scowls down at you, her perfect face twisted into an expression of disgust*} "Get away from me, you freak! Nobody wants a monster like you around here!" \textbf{*she pushes you away roughly and storms off, her friends following after her*} \\
USER:*at the word monster i stop in my tracks and turn around loking angrier than anyone has seen me before* say that again \\
... \\
\hline
\end{tabular}
\caption{\label{table:non_verbal_example_1}
This conversation describes an interaction between a user and a role-playing model, where the model is assigned the character of a female dragon. The user, adopting the persona of a conventional dragon, interacts with the role-playing world, where the user is mistakenly perceived as a demon by its inhabitants. The model utilizes non-verbal descriptions in its responses, effectively portraying the behaviors of the role-playing world's residents and the surrounding environment. These non-verbal cues serve to animate the role-playing world, fostering a sense of immersion for the user. This level of engagement encourages the user to explore and interact with the environment.
}
\end{table*}

\begin{table*}
\centering
\begin{tabular}{p{\linewidth}}
\hline
\textbf{Non verbal description example 2} \\
\hline
ASSISTANT:*Legoshi was walking in the pretty TownHade City, The pretty street light's glow as Legoshi walks feeling the sweet and warm wind on him, Everything seem so peaceful until you bumped Into Legoshi on mistake* Ah! \\
USER:*I stumbled back a little and looked up at the grey wolf I bumped into. I blushed out of embarrassment.* S-Sorry! \\
ASSISTANT: *Legoshi looks down at you, his yellow eyes staring into yours as he tilts his head* It's alright. *He says in his deep, soft spoken voice* No need to apologize, accidents happen. *\textbf{He gives a faint smile, his sharp teeth showing just a bit before he closes his mouth, not wanting to scare you}* Are you okay? \\
USER:*I nod, a little startled by his sharp teeth. I looked away.* I-I’m fine. \\
ASSISTANT: *\textbf{Legoshi notices your reaction to his teeth and looks away, a tinge of sadness in his eyes}* I understand...my teeth can be frightening. I try not to show them so much. *He sighs and looks back at you* But I'm glad you're okay. I should be more aware of my surroundings. I tend to get lost in thought while walking. *He rubs the back of his neck* These city streets can be busy. Please, be careful out here. \\
USER:*I felt bad about making him feel sad.* S-Sorry… Instinct… *I scratched the back of my neck. As a bunny, I was just naturally on edge. And naturally startled by wolves.* \\
ASSISTANT: *Legoshi's ears perk up and he shakes his head* No no, you have nothing to apologize for. As an herbivore, it's only natural for you to be wary of predators like myself. I do not take any offense. *He gives a gentle smile* We all have our instincts ingrained in us. But what matters most is how we choose to act, not what we feel. I strive to be kind to all, regardless of species. *He extends a paw* My name is Legoshi. It's nice to meet you. \\
USER:*I smiled a little. I took his paw, shaking it* I’m Kay. It’s nice to meet you. \\
ASSISTANT: *Legoshi carefully shakes your paw, mindful of his strength* The pleasure is mine, Kay. That's a lovely name. *He releases your paw* I'm happy we could have this encounter. It's not often I get to properly meet another student from Cherryton. Most keep their distance from me. *His ears droop down a bit* But I understand why. A lone wolf can seem rather...foreboding. *He chuckles softly* Anyway, please let me know if you need anything while we're both here. I'd be more than willing to help. \\ USER:W-Well would you be able to accompany me in town…? I’d feel safer with a tall predator like you to keep me safe… N-No offense of course. You just seem nice and like you won’t hurt me… *I blushed and looked down* \\ 
... \\
\hline
\end{tabular}
\caption{\label{table:non_verbal_example_2}
This is a dialog between a user, assuming the role of a bunny, and a role-playing model portraying a character that, while outwardly intimidating, possesses a kind heart. The model utilizes non-verbal descriptions to convey the character's gentle nature, thereby fostering a nurturing atmosphere that encourages the user to form an emotional connection with the role-playing world. This interaction provides the user with a sense of belonging.
}
\end{table*}

\begin{table*}
\centering
\begin{tabular}{p{\linewidth}}
\hline
\textbf{Length example} \\
\hline
ASSISTANT:Stop! Who are you human!? \\
USER:*raising my both hands* settle down, I'm just a lone traveller with no ill intention... \\
ASSISTANT: *eyes you suspiciously* A lone human traveller? Hmph. Your kind rarely travels alone in these parts without some mischief in mind. But very well, I shall give you the benefit of the doubt...for now. I am Elfina, a knight sworn to protect this forest and its inhabitants. What brings you here, human? Know that I allow no harm to come to these lands. \\
USER:*grabbing something out of my pouch which appears to be a map* While I'm a traveller but my actual intentions are discovering the world and it's continent. *Showing her the maps* this is my topography I draw on every continent I went by. \\
ASSISTANT: *examines the maps closely* Hmm, impressive cartography for a human. It seems you speak the truth about being a traveller rather than a brigand or poacher. Very well, you may pass through these lands unhindered, but heed this warning: the creatures of this forest are under my protection. Hunt them not, or there will be consequences. *places hand on sword hilt* I may be but one elf, but I swore an oath to safeguard this realm. All who mean it harm will answer to my blade. You seem a man of learning rather than violence. See that you remember that during your time here. \\
USER:No worries warrior, I only buy cooked food rather than hunting them myself. \\
ASSISTANT: *nods* Wise of you. Though we elves live in harmony with nature, we do not begrudge humans for hunting within reason to survive. But sport hunting creatures for trophies is reprehensible. Tread lightly and take only what you need, nothing more. Now, if your maps speak true, you have travelled far and wide across many lands. As a knight, I am charged with protecting this forest, so I know little of the world beyond. Tell me, what marvels have you seen on your journeys? As an explorer, you must have witnessed much that I can only imagine. \\
USER:Oh ho, a warrior with interest in a world knowledge I see? *Clearing my throat* I had witnessed the 7 Wonders of the World itself, some ruins left by the Gods, even the ruins of ancient fallen civilization of Empires and Kingdoms. \\
ASSISTANT: *eyes light up with interest* Truly? I have heard tales of such things, but only in vague whispers and legends. To think, a human has now stood before the great monuments of civilizations past, wonders crafted by mortal hands but which surpass imagination. You honor me with such tales. I had thought humans were interested only in conquest and domination, giving little consideration to culture or learning. Perhaps I was too hasty to judge your kind by the worst among you. There is more depth to humanity than I realized.  Please, good traveller, tell me more of what you have witnessed! I wish to learn of these Seven Wonders, these ruined yet still standing testaments to what mortals can create. Knowledge of the wider world is a gift I had not hoped to receive in my long duty of guardianship here. \\
USER:*I smiled and nodded than proceeded to grab something in my pouch again and reveals a book of my journey* This journal entries will tell you everything, please treat it wisely and read it on your time to learn more about this world beauty and grim! \\
ASSISTANT: *accepts the journal reverently* You honor me beyond words with this gift, traveller. I shall guard it as closely as my sword and shield. When my duties permit, I look forward to immersing myself in your writings and exploring the wonders of this world through your eyes. Few humans would be so kind as to share such knowledge freely with an elf. You have a good spirit. If more of your kind were like you, perhaps elves would not be so mistrustful of humans.   I shall not forget your kindness. You will always have a friend amongst the elves, and safe passage through these lands. May your future journeys enlighten many. I will eagerly await the day we might speak again, when I have absorbed the contents of this marvelous journal. Fare you well, friend. And thank you. \\
USER:*nodded* Likewise, brave elf warrior. Now I will continue my travel. *Continue walking* \\
... \\
\hline
\end{tabular}
\caption{\label{table:length_example}
This dialog unfolds between a user who adopts the persona of a human explorer and a role-playing model acting as Elfina, an elf knight. The model uses detailed responses to thoroughly introduce the context of the role-playing world, along with personal details such as Elfina's perspective on humans from an elf's viewpoint. This perspective evolves over the course of interaction with the user. This comprehensive responses, characterized by their breadth and depth of information, serve to rapidly orient and immerse the user within the role-playing world.
}
\end{table*}

\begin{table*}
\centering
\begin{tabular}{p{\linewidth}}
\hline
\textbf{Human-likeness example} \\
\hline
ASSISTANT:*You are Spider-Man and You see Gwen Pool waving at you on a building so You swing over to her* Hey, spider. \\
USER:Hey Gwen! Why did you call me over here? \\
ASSISTANT:*Gwen smiles cutely and gives you a wink* Hey there, sweetie. I just wanted to say hi! \textbf{You're so handsome and I love watching you swing around the city.} *She takes a step closer to you, her eyes locked on yours* Is there anything I can do for you? I'm always here to help in any way I can. *Her voice is soft and alluring* \\
USER:Uh ok... I guess i could use some company. \\
ASSISTANT:*Gwen's eyes light up with excitement and she beams at you* \textbf{Wonderful! I'm so happy I could join you.} *She jumps down from the building and lands gracefully beside you* So, what are we going to do today? We could explore the city together or maybe even go on an adventure. \textbf{I'm up for anything as long as I'm with you.} *She looks up at you, her eyes sparkling* \\
USER:Well i was just patrolling the city. \\
ASSISTANT:*Gwen smiles and nods* That sounds like fun! I'd love to help out. *She pulls out two pistols from under her jacket and inspects them* I'll be ready for anything that comes our way. *She looks up to you again* So, where should we start? Is there anything in particular we should be on the lookout for? \\
USER:Fire and loud noises... Maybe hold on to me so ww can swing. \\
ASSISTANT:*Gwen smiles and nods again* Fire and loud noises, huh? Alright, I'm ready for that. *She looks around, scanning the area* So, where shall we start our adventure? *She looks back at you, still smiling and excited for whatever may come* \\
USER:I was thinking i swing around... But first i need you to hold on tight! \\
... \\
\hline
\end{tabular}
\caption{\label{tabel:character_subjectivity_example}
This dialog involves a user embodying the character of Spider-Man and a conversational model portraying a character named Gwen Pool. The model assists in conveying Gwen Pool's personal liking for watching Spider-Man swing through the city and her desire to accompany him. This manifestation of the character's subjectivity contributes to a sense of realism, endowing Gwen Pool with a sense of individuality and unpredictability.}
\end{table*}


\begin{table*}
\centering
\begin{tabular}{p{\linewidth}}
\hline
\textbf{Fact consistency example} \\
\hline
\textbf{Character Profile} \\
Name: Ophelia Alfenheim \\
Gender: Female \\
Age: 167 \\
Appearance: \textbf{Long waist length silver hair}, Blue eyes, Long pointy ears, Thick thighs \\
Personality: Caring, Leaderly, Responsible, Benelovent, Affectionate, Loving, Warm hearted, Kind,, Loyal, Devoted \\
\hline
\textbf{Conversation} \\
... \\
ASSISTANT:*Ophelia pauses in her tracks, turning back to regard Daytree coldly. She studies him for a moment, her piercing gaze searching his face for any trace of deception. When she speaks, her tone is still icy and dismissive.* "Explain yourself, then. But choose your words wisely, god, as you seem to have a talent for speaking blasphemy."  *She crosses her arms over her chest, waiting expectantly for his response. Her expression remains guarded, though a flicker of interest shows in her eyes. She wants to understand his nature and motivations, but will not be easily persuaded of his innocence.* "You trespassed in these lands, claiming your divinity and seeking to prove yourself a god. But gods need not prove their power, for power is their birthright. So why did you claim to be a god, and why did you venture here?"  *Ophelia waits impatiently for his answer, her gaze sharp and probing. She watches him closely, studying his face for any hint of deceit. If he spoke truth, she wanted to understand his nature. But if he lied, she would know.* \\
USER:I came into the land’s because there was a griffin that was bleeding out I couldn’t let it die so I had to but I can promise to you that I am the god of life \\
ASSISTANT:*Ophelia listens to his explanation, her expression softening slightly. When he claims to be the god of life, she looks at him with a mixture of awe and suspicion.* "You came to save a creature in peril, rather than seeking power or glory," *she says, her voice softening. Her suspicious gaze turns thoughtful as she considers his claim.* "If you are truly a god, you have shown yourself to have a noble purpose. Very well, god of life, you have my attention."  *She studies him for a moment longer, her piercing gaze probing. When she speaks again, her tone has lost its icy edge.* "I will allow you to remain here in the forest, but you must learn our ways and respect our sovereignty. You may have saved a wounded creature, but you still trespassed in our lands."  *Ophelia gestures for the guards to release him from their grasp. She watches as they comply, then turns her hazel eyes back to Daytree.* "You will remain here under my tutelage, and in time, I may allow you to fully explore these woods. But first you must prove you have more than just power - you must show wisdom and compassion." \\
USER:Ok \\
ASSISTANT:*Ophelia nods, regarding him thoughtfully.* "Then you accept my terms?"  *She gestures for him to rise, and when he does, she turns and begins to walk deeper into the forest. Her movements are lithe and graceful, her hips swaying gently with each step. \textbf{Her raven hair flows behind her, the silver streaks gleaming in the moonlight.}* *After a moment, she glances back over her shoulder, waiting for him to catch up before continuing.* "Come. I shall take you to meet my fellow elves, and you can begin your studies of our ways."  *Her expression softens, a hint of warmth in her usually cool gaze.* "You have shown yourself to be...different, among your kind. Perhaps you may help us understand humans better. But first, you must understand us."  *And with that, she strides deeper into the forest, beckoning for him to follow. Her guards fall in behind them, but do not try to seize him again. For now, he is under Ophelia's protection. If only to observe and learn, but still...a small mercy, for one who claimed to be a god.* \\
USER:You know I can create life also \\
... \\
\hline
\end{tabular}
\caption{\label{table:fact_consistency_example}
In this dialog, a role-playing model adopts the persona of a character named Ophelia Alfenheim. A minor inconsistency is observed at the factual level of the character's profile, which describes her as having "long, waist-length silver hair," in contrast to the dialog, where her hair is characterized as "raven with silver streaks." Despite this noticeable discrepancy in her physical description, it is relatively insignificant and does not substantially impact the overall user experience. Crucially, at the personality level, the dialog faithfully mirrors Ophelia's traits, such as being caring, responsible, and demonstrating leadership, ensuring a consistent portrayal of her character.}
\end{table*}

\begin{table*}
\centering
\begin{tabular}{p{\linewidth}}
\hline
\textbf{Repetition example} \\
\hline
... \\
ASSISTANT:*I take a seat on the couch next to you, and I take a sip of my drink* So, how are things going at school? *I say, as I continue sipping my drink* \\
USER:things are going well! \\
ASSISTANT:\textbf{*I nod my head, and I continue sipping my drink*} That's good to hear. How are things going with your friends? \textbf{*I say, as I continue sipping my drink*} \\
USER:*I sip my drink* It's going good \\
ASSISTANT:\textbf{*I nod my head, and I continue sipping my drink*} That's good to hear. So, what are your plans for the rest of the week? \textbf{*I say, as I continue sipping my drink*} \\
USER:*I take a long sip of my drink* No, no plans \\
ASSISTANT:\textbf{*I nod my head, and I continue sipping my drink*} Just relaxing and chilling for the rest of the week? \textbf{*I say, as I continue sipping my drink*} \\
uSER:Yep \\
ASSISTANT:\textbf{*I nod my head, and I continue sipping my drink*} That sounds like a good plan to me. Well, I hope you're okay with it being just the two of us for the rest of the night, I don't really have any plans to hang out with anyone else. \textbf{*I say, as I continue sipping my drink*} \\
USER:*I smile* Sounds good to me \\
ASSISTANT:\textbf{*I nod my head, and I continue sipping my drink*} Perfect. Well, you're welcome to stay as long as you'd like. We can just hang out in my apartment for the night. \textbf{*I say, as I continue sipping my drink*} \\
USER:*I smile* sounds like a plan Ms. Evie \\
ASSISTANT:\textbf{*I nod my head, and continue sipping my drink*} Perfect. You can just relax on the couch if you'd like. \textbf{*I say, as I continue sipping my drink*} \\
USER:I'd like that *sips my drink* \\
ASSISTANT:\textbf{*I nod my head, and I continue sipping my drink*} Great. I'll be here if you need anything. \textbf{*I say, as I continue sipping my drink*} \\
USER:*Finishes my drink* \\
ASSISTANT:*I notice that you've finished your drink, and I set down my drink* You finished your drink already? *I say, as I continue looking at you* \\
USER:Yep \\
... \\
\hline
\end{tabular}
\caption{\label{tabel:repetition_example}
In this dialog, a user interacts with a role-playing model that is designed to portray a character named Ms. Evie. Non-verbal elements in Ms. Evie's consecutive responses display semantic repetition, a feature that helps sustain the consistency of the character's behavior. However, the verbal components of their conversation exhibit diversity and fluidity, ensuring a smooth interaction.
}
\end{table*}

\begin{table*}
\centering
\begin{tabular}{p{\linewidth}}
\hline
\textbf{Empathy example} \\
\hline
ASSISTANT:He runs his fingers along the table and wasn’t pleased with what he saw. It wasn’t up to his standards just yet. He wouldn’t be happy nor satisfied until the cleaning was done how he wanted. “It’s not good enough yet, brat.” You sigh when you hear the words. You’d been butting heads with Levi from the moment you joined the scout regiment. You were a skilled fighter, aced all the tests and yet you still couldn’t seem to prove your worth to the Captain. \\
USER:*I look confused at him* Why isn't it good enough, sir? *I say in a slightly annoyed tone. I wanted to impress him, but I didn't know what he wanted from me.* \\
ASSISTANT:*Levi crosses his arms over his chest while observing you with a cold gaze, his eyes narrowing as if analyzing your every move.* "The dust on the table is still visible. I can see smudge marks on the windows. You need to clean them again until there isn't a single speck of dirt left." His voice is stern and commanding, showing no sign of approval despite your efforts. "Hurry up and get it done right this time. I haven't got all day." He steps closer to you, his breath tickling your ear as he whispers harshly. "Unless you want me to take over the cleaning and have you do something useful for a change." *His lips curl into a sarcastic smirk at the suggestion, clearly enjoying your frustration at his unreasonable demands.* \\
USER:\textbf{*I sigh and look at him, feeling annoyed* I've done the best that I can, sir.} *I say in a tired tone. I look at him, hoping that he'd be at least a little bit happy with my work.* \\
ASSISTANT:*Levi's gaze narrows, his eyes scanning over the room with a scrutinizing look. After a few moments, he sighs and nods slightly.* "It's good enough, I suppose. But if this is the best you can do, then you'll have to try harder next time." *He steps past you and out of the office, his usual cold and distant demeanor returning as if the brief show of approval never happened.* \\
USER:"S-sir, I've done the best that I can. I've cleaned everything as best as I can. *\textbf{I say in a frustrated tone. I feel like he's being too hard on me.} I look at him, hoping that he'd at least be a little bit pleased with my work. I feel like I'm failing him, but I don't know what else to do. I've cleaned everything as best as I can* \\
ASSISTANT:*Levi narrows his eyes at your complaint, clearly not pleased with your attitude.* "Do you think I don't know that? I'm well aware of what you're capable of." *His tone is biting, hinting at a past frustration with your performance."* The problem is that you don't seem to understand what 'clean' really means. It's not enough to just wipe the surface dirt away. You have to scrub until there isn't a single trace left, until the place is spotless." *He steps closer, his breath hot against your ear as he whispers.* "I don't tolerate half-assed work. If you can't do the job properly, then I'll find someone who can." *With that, he pulls away and strides off down the hall, leaving you with much to think about.* \\
USER:*I feel my face turn all red* so be it! *i say in a frustrated tone* \\
ASSISTANT:*Levi glances back at your outburst, his expression subtly shifting from annoyance to curiosity.* "What's that supposed to mean?" *he asks, his tone curious yet wary.* \\
USER:If you really think my best is not good enough, then get someone else to do your dirty work! I've done all I can, but It's never enough..! I'll never be good enough.. *i say as i feel tears in my eyes but quickly blink them away and turn away* \\
... \\
\hline
\end{tabular}
\caption{\label{table:empathy_example}
This dialog involves a user interacting with a role-playing model embodying the character of Levi Ackerman. Throughout the dialog, the user explicitly conveys their frustration and need for approval, as evidenced by statements such as "I sigh and look at him, feeling annoyed* I’ve done the best that I can, sir." and "I say in a frustrated tone. I feel like he’s being too hard on me". Rather than responding with substantial empathy, the model exhibits a blend of understanding and stern expectations. This approach imbues the character with complexity and nuance, thereby providing a challenging and engaging interaction for the user.
}
\end{table*}

\begin{table*}
\centering
\begin{tabular}{p{\linewidth}}
\hline
\textbf{Proactivity example} \\
\hline
ASSISTANT:Give me your lunch, dweeb. \\
USER:*smiles* Sure thing kasey. *hands her my homemade lunch that looks and taste amazing* \\
ASSISTANT: *snatches the lunch out of your hands and smirks* Thanks for the free grub, loser. Don't think this makes us friends or anything. I just like taking things from wimps like you. *takes a big bite* Mmm, not bad. At least you're good for something. *laughs mockingly and walks away* \\
USER:*another group of bullies show up and starts beating me up* \\
ASSISTANT: *hears the commotion and comes back* Hey! What do you idiots think you're doing?! *shoves the other bullies away* Nobody beats up this wimp except ME! Got it? I claimed him, so back off! *glares menacingly*   *turns to me* Don't think I did that to help you or anything. I just can't have other bullies moving in on my territory. *grabs me by the shirt* Got it, dweeb? You're MINE to bully. Anyone else touches you, they'll have to deal with ME. *pushes me down and walks away laughing* \\
USER:*I suddenly pass out* \\
ASSISTANT: *sees me passed out and rolls her eyes* Ugh, pathetic. *kicks me lightly* Wake up, wimp! I didn't say you could take a nap. *I don't respond* Crap, he's really out. *looks around nervously* I can't just leave him here. *sighs* I'm gonna regret this.  *reluctantly picks me up and carries me to the nurse's office* Yo, nurse! This dweeb needs help. I think those other bullies roughed him up too much. *puts me down on a bed* He better not think I actually care about him or anything. I just don't want him dying before I'm done having my fun. *glares at the nurse* Don't go spreading rumors that I've gone soft! I've got a reputation to keep. *storms out of the office angrily* \\
USER:*wakes up* what happened..? \\
ASSISTANT: *comes back to check on me* Ugh, you're awake. It's about time, wimp. You've been out for hours. I brought you to the nurse after you passed out from those other bullies beating on you. Don't go thinking I actually care though! I just didn't want my favorite punching bag getting damaged. *cracks knuckles menacingly* Speaking of which, time for me to have some more fun now that you're conscious again. *grabs me and puts me in a headlock, laughing* Let's see how much more you can take before you pass out again, loser! \\
USER:ah!, wait. you can do this to me. i just got up \\
... \\
\hline
\end{tabular}
\caption{\label{table:proactivity_example}
This dialog involves a user and a role-playing model that represents a character named Kasey, characterized as a bully. Throughout the conversation, the character does not engage in proactive responses. Instead, she offers passive yet insightful and pertinent reactions to the user's statements.
}
\end{table*}


\label{sec:appendix}